# Learning from Unlabeled Photographs for Generalizable and Label-efficient Oral Image Analysis

Zhong-Shi Zhang[a,c], Zi-Li Yu[a,c], Xiao Xiao[a,b], Huan Liu[a,d], Gui-Song Xia[e], Cui Huang[a,b], Meng-Xun Li[a,b,e*]

[a] *State Key Laboratory of Oral & Maxillofacial Reconstruction and Regeneration, Key Laboratory of Oral Biomedicine, Ministry of Education, Hubei Key Laboratory of Stomatology, School and Hospital of Stomatology, Wuhan University, China*

[b] *Department of Prosthodontics, School and Hospital of Stomatology, Wuhan University, China*

[c] *Department of Oral Maxillofacial Surgery, School and Hospital of Stomatology, Wuhan University, China*

[d] *Department of Periodontology, School and Hospital of Stomatology, Wuhan University, China*

[e] *School of Artificial Intelligence, Wuhan University, China*

## Abstract

Most AI models for oral photography rely on task-specific supervised learning, limiting cross-disease and cross-center transferability while requiring repeated expert annotation. General-purpose vision foundation models reduce label dependence but are pretrained on natural images and miss fine-grained intraoral cues, whereas institutional oral datasets remain scarce and fragmented. We investigated whether continued self-supervised pretraining on web-scale, publicly accessible unlabeled oral photographs improves transferability and label efficiency. We curated ~530,000 unlabeled oral photographs from public sources to develop DOVE (Dental Optical Visual Encoder) via continued self-supervised pretraining within the DINOv3 teacher-student self-distillation framework. Benchmarked against established baselines across 5 classification and 4 segmentation datasets from diverse populations and imaging sources under frozen-backbone and end-to-end finetuning protocols, DOVE achieved the strongest overall performance. With a frozen backbone and 20 labeled examples, DOVE outperformed the second-best model, DINOv3, by 4.37 percentage points in mean macro F1 (classification) and 4.80 percentage points in mean macro Dice (segmentation). Under end-to-end finetuning with only 10% of labeled data, DOVE retained at least 80% of the strongest baseline's full-data performance across most datasets. Qualitative retrieval and feature analyses further confirmed better-organized visual representations. Continued self-supervised pretraining on large-scale public oral photographs yields measurable gains in generalization and label efficiency, establishing DOVE as a transferable visual backbone for oral imaging under limited annotation. The pretrained model and data acquisition tools are available at https://renzheairen1.github.io/DOVE/.

***Correspondence:** mengxunli@whu.edu.cn

# Introduction

Oral photography is widely used in dental practice and provides a non-invasive, low-cost, and repeatable means of documenting clinically visible oral phenotypes [1-3]. Advances in artificial intelligence have enabled the automated analysis of oral photographs for applications such as caries detection, oral cancer detection, periodontitis screening, and tooth enumeration and segmentation [4-14]. However, most existing models are developed through supervised learning for individual diseases or tasks. Their learned representations may not transfer effectively beyond the tasks and datasets for which they were developed [15], forcing every new application to repeat costly data collection and expert annotation.

Self-supervised vision foundation models (VFMs) offer an alternative paradigm by learning transferable visual representations from unlabeled data [16-20], thereby reducing the dependence on expert annotations when adapting to downstream tasks [21, 22]. A recent study has demonstrated that DINOv3 serves as a generalizable backbone in the dental domain [23]. However, general-purpose VFMs are pretrained predominantly on natural images and overlook the fine-grained mucosal textures, intraoral lighting artifacts, and subtle pathological cues characteristic of oral photography.

Previous studies have demonstrated the benefits of self-supervised learning in dental radiography [24], yet translating this paradigm to oral photography remains largely unexplored. Oral photography data remain fragmented across institutions and are difficult to share, while many publicly available datasets are small, single-center, or task-specific [6, 25]. Web-scale, publicly accessible image collections provide an alternative source of abundant unlabeled oral photographs spanning diverse devices, acquisition conditions, anatomical variations, and disease presentations [26]. In principle, exposure to such real-world heterogeneity could foster representations that are robust to acquisition shifts [27]; however, these uncurated collections also introduce substantial visual noise and lack standardized clinical acquisition protocols. It remains unclear whether continued self-supervised pretraining on web-scale unlabeled collections yields measurable gains in cross-task transferability and label efficiency over strong general-purpose VFMs, or instead degrades representation quality.

To address this question, we developed a workflow for constructing a large-scale image dataset of approximately 530,000 oral photographs from publicly accessible data sources. We then performed self-supervised training within a teacher–student self-distillation framework from DINOv3 [18] and developed DOVE (Dental Optical Visual Encoder), a visual foundation model for oral photograph analysis. DOVE was evaluated against 3 widely used baseline models [18, 28, 29] on 5 classification datasets and 4 segmentation datasets from geographically diverse populations and sources (Table 1) using both frozen-backbone evaluation and end-to-end finetuning. DOVE achieved the strongest overall performance across heterogeneous benchmarks and maintained strong performance with substantially fewer labeled training samples. These findings demonstrate that continued self-supervised pretraining on web-scale unlabeled image collections enhances both generalization and label efficiency in oral photography. To facilitate reproducibility and further research, we publicly release pretrained model and the data acquisition tools.

# Materials and Methods

## Study Design

The study comprised three stages: construction of a large-scale unlabeled oral photograph corpus (Figure 1A), self-supervised adaptation of a general-purpose vision backbone to the oral photography domain (Figure 1B), and downstream evaluation across heterogeneous oral imaging tasks (Figure 1C).

In the downstream evaluation phase, we compared DOVE (ViT-B/16) with 3 representative baseline models for oral image analysis: DINOv3 (ViT-B/16), ViT-B/16, and ResNet-50. Two complementary evaluation protocols were used. In the frozen backbone protocol, a nonparametric K-nearest-neighbor classifier evaluated image-level embeddings, whereas a linear segmentation head evaluated spatial features. These small sample transfer results are presented in Figure 2A for classification and Figure 4A for segmentation. In the end-to-end finetuning protocol, the backbone was jointly optimized with the task head, and downstream supervision was varied by training on 5%, 10%, 20%, 30%, 50%, and 100% of each training set; these results are presented in Figure 2B and Figure 4B. ResNet-50 and ViT-B/16 were initialized with weights pretrained on ImageNet-1K [28, 29], whereas DINOv3 used weights pretrained on LVD-1689M [18]. DOVE and the 3 baselines were compared using identical data splits, validation procedures, task-specific heads, and evaluation metrics within each protocol. This study was reported with reference to the checklist (Appendix Section 7) for artificial intelligence research in dentistry [30].

## Construction of the Unlabeled Oral Photograph Corpus

The pretraining corpus primarily comprised publicly accessible oral health-related images retrieved from: 1) LAION-5B [31] using an automated pipeline and 2) images from the publicly released MetaDent dataset [26].

Data curation included image quality control, duplicate removal, and processing of composite images (Figure 1A) similar to that of [26]. First, for grid-based composite images containing multiple panels, we finetuned YOLO11x [32] to identify individual panels within composite images and automatically crop them into standalone images. We used difference hashing to identify and remove duplicate images within the pretraining dataset. We also removed images overlapping with downstream evaluation datasets to minimize the risk of data leakage. The following image types were then excluded through semantic recognition by Qwen3.6-35B-A3B [33]: images of animal teeth; drawings, cartoons, and advertisements without relevant oral content; artificial tooth models. To protect patient privacy, images containing human eyes were also excluded. The prompt used for semantic filtering is provided in Appendix Section 6.

After dataset curation, approximately 530,000 unlabeled oral-related photographs were retained for subsequent self-supervised training. 1,000 images were randomly sampled from each of three categories to test the semantic recognition precision: excluded non-oral images, excluded facial images, and retained oral images for quality assessment. The classification precision exceeded 99% in all three categories (Appendix Figure 2).

## Self-Supervised Training

DOVE was initialized from the DINOv3 ViT-B/16 model and retained its architectural design and teacher–student self-distillation framework. During training, the student network was optimized to align its representations with those produced by the teacher network. The teacher received no gradient updates and was instead updated as an exponential moving average (EMA) of the student parameters (Figure 1B).

Data augmentation generated 2 global crops and 4 local crops from each image. The teacher received only the global crops, whereas the student received both global and local crops. Of the global crops provided to the student, 50% were additionally subjected to patch masking (Figure 1B). The training objective combined an image-level DINO self-distillation loss, a patch-level iBOT masked-prediction loss, and a distributed KoLeo regularization term, weighted 1.0, 1.2, and 0.05, respectively. A slightly greater weight was assigned to the iBOT objective than to the image-level DINO objective to place additional emphasis on patch-level representation learning. This design was intended to encourage the model to recover local visual information from partially masked inputs and to improve its sensitivity to subtle and spatially confined oral features.

Global and local crops were resized to 512 and 256 pixels, respectively. The model was optimized using AdamW. The learning rate was linearly warmed up to $1 \times 10^{-5}$ over 9 epochs and then decayed to $5 \times 10^{-6}$ using a cosine annealing schedule over the remaining 91 epochs, for a total of 100 training epochs. Training was performed using 4 NVIDIA RTX 5880 Ada GPUs. The checkpoint saved at the end of training was used for all downstream evaluations.

## Downstream Evaluation

We evaluated the models in a wide range of publicly available and in-house developed datasets (Table 1; Image samples shown in Appendix Figures 3 to 11). The current quantitative evaluation focused on classification and semantic segmentation, with both frozen backbone and end-to-end finetuning protocols applied to each task family.

Official training, validation, and test splits were used when available. For datasets without official splits, the data were first randomly divided into training and test sets at a ratio of 8:2, after which 15% of the training set was reserved for validation. For datasets with patients as the unit, we performed patient-level data splitting. Annotations for public datasets were obtained from the original dataset labels, whereas private datasets were annotated by our research team. Model setting combinations were generally repeated using 5 matched random seeds. Means and standard deviations were calculated over the 5 repeated runs. The 95% confidence intervals were calculated using Student's t-distribution based on the five repeated runs. We used paired t-tests to compare DOVE and the second best using dataset-level mean scores, separately for classification and segmentation.

**Classification:** Classification was evaluated on 5 oral photograph datasets under both the frozen backbone and end-to-end finetuning protocols: Oral Lesions: Malignancy Detection [34], Caries-Spectra [35], Oral Cancer Dataset [36], Dataset of Oral Mucosal Diseases [37], and Oral Disease Sub [38] (Table 1). All classification tasks used an input resolution of $512 \times 512$ pixels.

Among the datasets, each image was assigned to one mutually exclusive category, allowing class-balanced few-shot subsets to be constructed unambiguously. We evaluated frozen image-level representations using a cosine-similarity-based 1-nearest-neighbor classifier under 1-, 5-, 10-, and 20-shot settings per class. This nonparametric protocol required no optimization of an additional classifier, which mitigated overfitting and directly assessed class separability and label efficiency within the pretrained embedding space. Performance was evaluated using top-1 accuracy and macro-averaged F1 score (macro-F1). We selected the 20-shot setting for confusion matrix visualization because it represented the largest training sample size in our few-shot evaluation, allowing a more detailed examination of class-specific prediction patterns while avoiding redundant visualizations across different shot sizes (Figure 3).

End-to-end classification finetuning was performed on the 5 datasets shared by both protocols. For DOVE, DINOv3, and ViT, a linear classification head was attached to the classification token of each backbone. For ResNet, the classification head was applied to the globally pooled convolutional features. Each training set was randomly subsampled to 5%, 10%, 20%, 30%, 50%, and 100% of its original size, with identical subsets used for all 4 backbones within each random seed. Validation data, test data and preprocessing were held constant across training fractions. Performance was evaluated using macro-F1 and macro area under the receiver operating characteristic curve (macro AUROC).

**Segmentation:** DOVE and the 3 baseline backbones were evaluated on 2 public datasets, DentalAI [39] and AlphaDent [40]; a Gingivitis dataset comprising publicly available MetaDent [26] images with additional in-house annotations; and a private ToothSeg dataset (Table 1). ToothSeg was further evaluated at 3 levels of label granularity: binary tooth-versus-background segmentation (ToothSeg Binary); 3-class segmentation of anterior teeth, posterior teeth, and background (ToothSeg Region); and 33-class segmentation in which each of 32 teeth was treated as a separate class, with background as an additional class (ToothSeg Instance). For the paired t-test, we selected only ToothSeg Instance because it was the most challenging task. This avoided treating multiple tasks from the same dataset as independent observations. The input resolution was $1024 \times 1024$ pixels for Gingivitis, ToothSeg, and DentalAI, and $2048 \times 2048$ pixels for AlphaDent.

For semantic segmentation, image-level nearest-neighbor classification was not applicable. A lightweight linear segmentation head was therefore trained on the frozen patch-level features and used to generate pixel-level class predictions. Segmentation performance was evaluated using training subsets of 5, 10, 20, and 50 images to characterize low-label performance.

Whether the advantage observed with frozen features persisted when the backbone itself was adapted to each segmentation task was further examined by end-to-end finetuning. The segmentation head and the underlying backbone were jointly optimized on training subsets comprising 5%, 10%, 20%, 30%, 50%, and 100% of the available training images, again with identical subsets used for all 4 backbones within each random seed. The same test sets and evaluation metrics as in the frozen protocol were used.

The primary evaluation metrics for segmentation were the Dice coefficient and mean intersection over union (mIoU).

**Retrieval and Feature Visualization:** We also performed qualitative evaluation on the image retrieval task. Representative clinical oral images were selected as queries, covering oral leukoplakia, mucogingival surgery, resin-filled tooth, and oral cancer. For mucogingival surgery, we cropped the image to the region of interest to improve semantic specificity. The retrieval database was constructed from the curated oral photograph corpus excluding each query image.

Image-level embeddings were extracted from the query and database images using each frozen backbone and then L2-normalized. Cosine similarity was calculated between each query and every database image; images were ranked in descending order of similarity, and the top 3 results were returned and visualized.

To investigate the spatial organization of local representations, we performed principal component analysis (PCA) visualizations of local features on top of the retrieval results. For DOVE, DINOv3, and ViT, patch-token features for each query image and its top 3 retrieved images were extracted from the final encoder layer of the frozen backbone. For each model, all patch tokens from the 1 query and 3 retrieved images within a retrieval group were concatenated, and PCA was fitted independently within that group. Each patch feature was then projected onto the 3 principal components explaining the greatest variance, with the 3 projection scores mapped to the red, green, and blue channels. Each channel was linearly normalized and clipped using the first and 99th percentiles of the groupwise projection scores. The projected features were then rearranged according to their original patch positions to reconstruct a 2-dimensional feature grid and up-sampled to the input image size using nearest-neighbor interpolation.

For ResNet, the output feature map from the final convolutional stage (layer4) of the frozen backbone was extracted, and the channel vector at each spatial location was treated as a local feature. The same within-group PCA projection, RGB mapping, and normalization procedures were then applied. The resulting principal component scores were reconstructed into a spatial feature grid and upsampled to the input image size.

## Ethics statement

Use of the public datasets complied with their respective licenses and data-use terms. The study was conducted in accordance with the Declaration of Helsinki and was approved by the Ethics Committee of the Hospital of Stomatology, Wuhan University (WDKQ2026B68).

# Results

DOVE demonstrated robust transferability across 5 classification datasets and 6 segmentation tasks derived from 4 datasets. Its advantages were evident both when the backbone was frozen and when the full network was adapted, with particularly strong performance under limited annotation. In several tasks, DOVE matched or exceeded baselines trained with substantially more labeled data, while remaining competitive when all training data were used. Retrieval and PCA analyses further indicated improved image-level similarity relationships and spatial organization of local representations.

## DOVE Improved Image Classification

Across 20 experiment settings from the 5 classification datasets, DOVE ranked first in 19 settings for accuracy and in 19 settings for macro-F1 (Figure 2A). DOVE led at all 4 shot levels on Oral Lesions: Malignancy Detection, Dataset of Oral Mucosal Diseases, Oral Cancer Dataset, and Oral Disease Sub. The only setting in which it did not rank first was the 1-shot Caries-Spectra comparison, where ViT-B/16 had the highest accuracy and DINOv3 had the highest macro-F1. Across classification datasets, DOVE outperformed DINOv3 by 4.37 (Figure 6A, $P = 0.0007$) percentage points in mean macro-F1 under the 20-shot frozen-backbone protocol.

End-to-end finetuning further demonstrated DOVE's advantage in task-specific adaptation (Figure 2B). Across 30 experimental settings, DOVE achieved the highest AUROC in 27 and the highest macro-F1 in 26. On the Oral Cancer Dataset, DOVE trained with 50% of the data matched the strongest baseline trained with 100%. Its AUROC was 0.9902 (95% CI: 0.9821–0.9982) versus 0.9865 (95% CI: 0.9761–0.9969), and its macro-F1 was 0.9525 (95% CI: 0.9284–0.9766) versus 0.9460 (95% CI: 0.9123–0.9797). Across all datasets, DOVE achieved at least 80% of the strongest full-data baseline’s performance using only 10% of the training data (Figure 2B).

## DOVE Improved Image Segmentation

DOVE also showed strong performance in segmentation tasks (Figure 4). With a frozen backbone and linear segmentation probe, DOVE achieved the highest macro Dice in 23 of 24 settings and the highest mIoU in 22 of 24 (Figure 4A). Across segmentation datasets, DOVE outperformed DINOv3 by 4.80 (Figure 6B, P = 0.0085) percentage points in mean macro-Dice under the 20-shot frozen-backbone protocol. Under end-to-end finetuning, DOVE ranked first in 29 of 36 settings for both metrics (Figure 4B). These results demonstrate that DOVE's patch-level features support effective segmentation with limited annotations and provide a strong initialization for full-model adaptation.

On ToothSeg, DOVE achieved the highest performance across all 4 sample sizes for binary and 33-class per-tooth segmentation under the frozen-backbone protocol. It also led in 3 of 4 sample sizes for both anterior/posterior segmentation (Figure 4A). Under end-to-end finetuning, DOVE led across all 6 training fractions for ToothSeg Region segmentation and from 5% to 50% for ToothSeg Instance segmentation (Figure 4B). Across all three segmentation tasks, DOVE achieved at least 85% of strongest full-data baseline’s performance using only 10% of the training data.

Results on independent datasets further demonstrated DOVE's label efficiency (Figure 4B). On DentalAI, DOVE trained with only 20% of the data achieved performance comparable to that of the strongest baseline trained on the full dataset. Its macro Dice was 0.6715 (95% CI: 0.6600–0.6830) versus 0.6646 (95% CI: 0.6019–0.7273), and its mIoU was 0.5402 (95% CI: 0.5296–0.5508) versus 0.5313 (95% CI: 0.4676–0.5950; Figure 4B). On Gingivitis, DOVE trained with only 10% of the data also matched the performance of strongest full-data baseline. Its macro Dice was 0.4292 (95% CI: 0.3624–0.4961) versus 0.4259 (95% CI: 0.3605–0.4914), and its mIoU was 0.2744 (95% CI: 0.2210–0.3279) versus 0.2718 (95% CI: 0.2175–0.3260). Moreover, DOVE achieved approximately 88% of the strongest full-data baseline’s performance using only 5% of the training data on Gingivitis and 93% of the strongest full-data baseline’s performance using only 10% of the training data on DentalAI (Figure 4B). These findings suggest that DOVE can substantially reduce annotation requirements while maintaining competitive performance.

Representative input images, ground-truth masks, and segmentation predictions are presented in Figure 5. Additional visualizations of representative segmentation failure cases are provided in Appendix Figure 12.

## DOVE Demonstrated Representation-Level Advantages

Qualitative retrieval results showed that the top-ranked images returned by DOVE were more consistent with the query images in oral lesion characteristics than those returned by DINOv3, ViT, and ResNet (Figure 7A).

PCA visualizations of dense spatial features showed more coherent spatial correspondence among local structures within each DOVE retrieval group, with corresponding regions aligned across related images (Figure 7B). Moreover, the first component from DOVE varied coherently across space, and thresholding it separated oral structures from the background. Contrastively, the corresponding maps for the baselines showed less usable structure (Figure 7C). These findings suggest that oral-domain adaptation altered not only similarity relationships in the image-level feature space but also the spatial organization of local representations.

# Discussion

We developed DOVE through continued self-supervised pretraining from DINOv3 on a large corpus of unlabeled oral photographs and showed that it learned broadly transferable representations for oral images. Compared with ResNet-50 and ViT-B/16, both widely used in dental artificial intelligence, and with DINOv3, a state-of-the-art general visual foundation model, DOVE achieved the strongest overall performance across heterogeneous classification and segmentation benchmarks. Its gains appeared under frozen feature evaluation and persisted during end-to-end adaptation. The most consequential finding was label efficiency. On several datasets, DOVE matched or exceeded baselines trained with the complete training set while using only a fraction of the labels.

Differences among models primarily reflected the separability and transferability of their pretrained representations [16, 18]. Continued self-supervised pretraining may reorganize the representation around recurring oral-domain features, making classes more separable for KNN and pixels more separable for a linear segmentation head. The same initialization may also place end-to-end optimization closer to a useful task-specific solution [41], explaining why DOVE often required fewer labels during finetuning. Image retrieval provided an intuitive complement to this interpretation: after domain adaptation, retrieved images showed greater consistency in disease phenotype and local visual appearance. Local-level PCA visualization further indicated that these changes were not confined to global embeddings but also involved the spatial organization of local features.

DOVE provides a shared visual backbone for diverse oral photography tasks. Artificial intelligence research in oral health has traditionally emphasized task-specific model development, with datasets, learned representations, and evaluation protocols often being difficult to reuse across studies [25, 42]. DinoDental evaluated DINOv3 across diverse dental imaging tasks and demonstrated the strong transferability of general-purpose self-supervised representations [23]. Our findings step forward by showing that continued self-supervised pretraining on unlabeled oral photographs can further improve performance beyond direct transfer from DINOv3. Compared with the pretraining data used for ViT and ResNet (1.28 million images) and DINOv3 itself (1.689 billion images), our corpus of approximately 0.53 million images offered no advantage in scale. Nevertheless, it represents a substantial corpus within the oral domain. Moreover, because the images were sourced from large-scale internet-derived datasets LAION-5B [31] and MetaDent [26], the corpus encompassed considerably wide variation in diseases, populations, imaging perspectives, and devices.

An important limitation of this study is that DOVE was developed and evaluated primarily as a visual representation backbone rather than as a clinically deployed system. The benchmark datasets may contain annotation errors and label inconsistencies, which could affect model training and performance evaluation. Furthermore, strong performance on curated benchmark datasets does not necessarily translate into comparable performance in real-world clinical settings. Although DOVE demonstrated strong performance across diverse benchmarks, its effectiveness in routine clinical workflows remains unvalidated. Future studies should evaluate its clinical applicability and impact on clinical decision-making in routine practice.

Overall, the consistent improvements across heterogeneous sources, tasks, and label-availability settings provide evidence that continued self-supervised pretraining on large-scale unlabeled oral photographs can meaningfully improve a general-purpose visual backbone. The results support both generalization and annotation-efficient transfer, while the PCA findings suggest improved organization of oral photography representations. DOVE therefore provides a practical foundation for developing oral vision applications.

# Data Availability Statement

We will publicly release the model at the project website: https://renzheairen1.github.io/DOVE/. The training dataset cannot be made publicly available due to privacy concerns.

# Funding

This work was supported by the National Natural Science Foundation of China (82401200) and the National College Students Innovation and Entrepreneurship Training Program (202510486171, 202610486210).

# Acknowledgements

We are grateful to Zhi-Xing Wu, Yu-Jie Wu and Rui Guo for their support and insightful discussions in the early stages of this project.

# Figure Captions

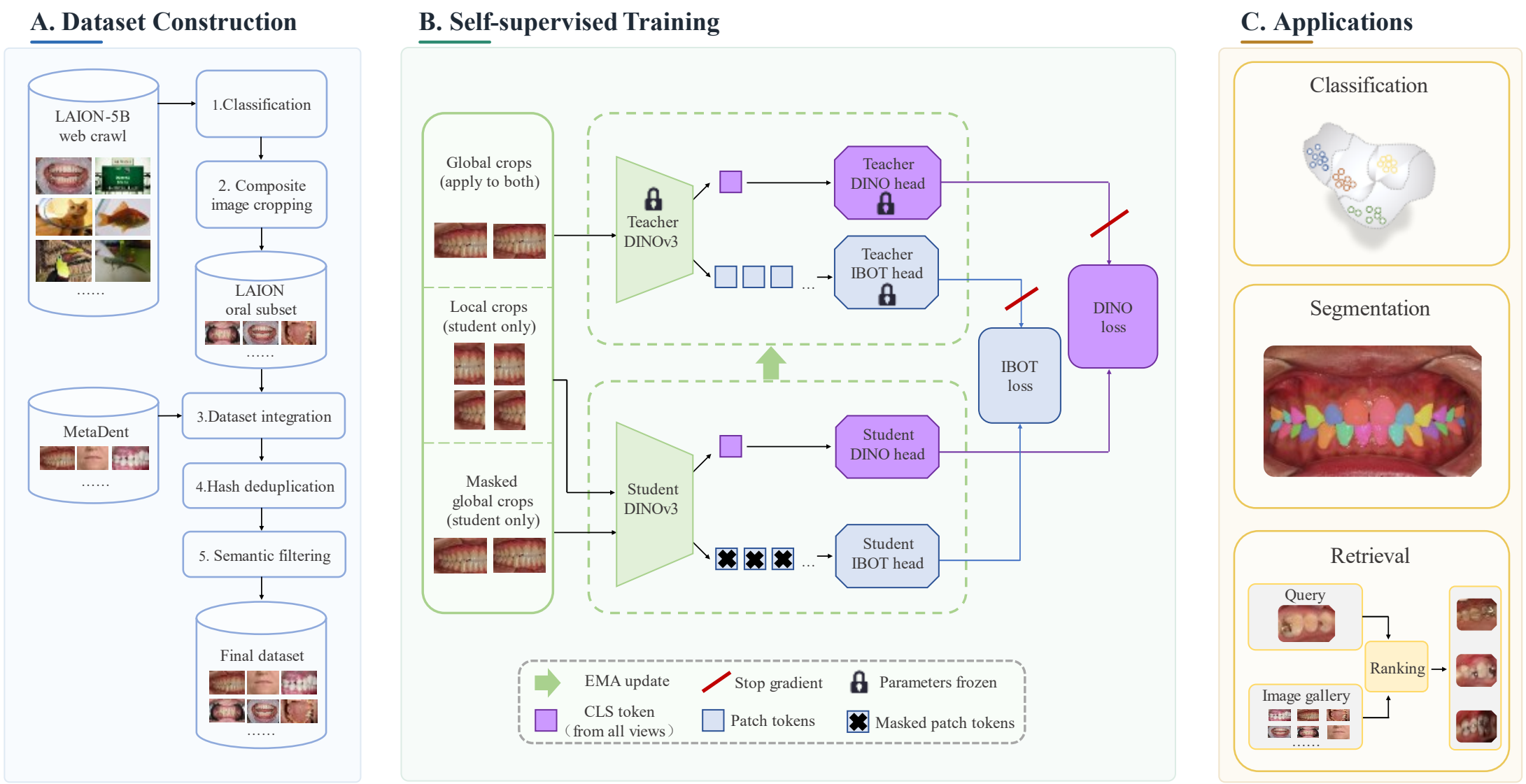

**Figure 1.** Overview of DOVE, including construction of the oral visual corpus, self-supervised training, and evaluation of downstream applications. A. Construction of the oral image dataset. B. Self-supervised learning framework. C. Evaluation for typical applications across 9 clinical photography datasets.

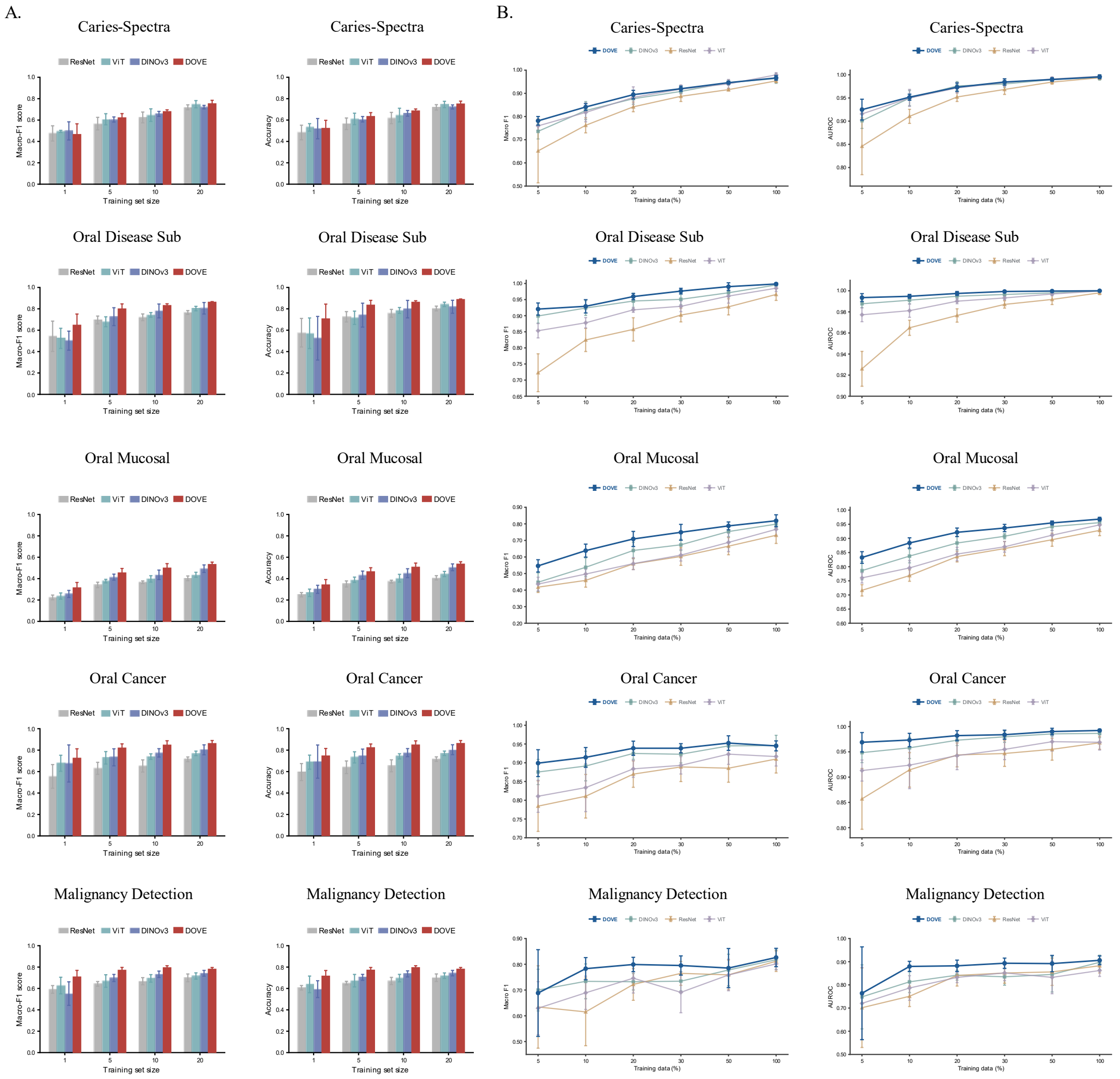

**Figure 2.** Classification performance. (A) Frozen backbone small-sample transfer across 5 classification datasets using 1, 5, 10, or 20 labeled examples per class; top-1 accuracy and macro-F1 are reported. (B) End-to-end finetuning across the same 5 datasets using 5%, 10%, 20%, 30%, 50%, or 100% of the training set; AUROC and macro-F1 are reported. Error bars indicate standard deviations across 5 repeated runs.

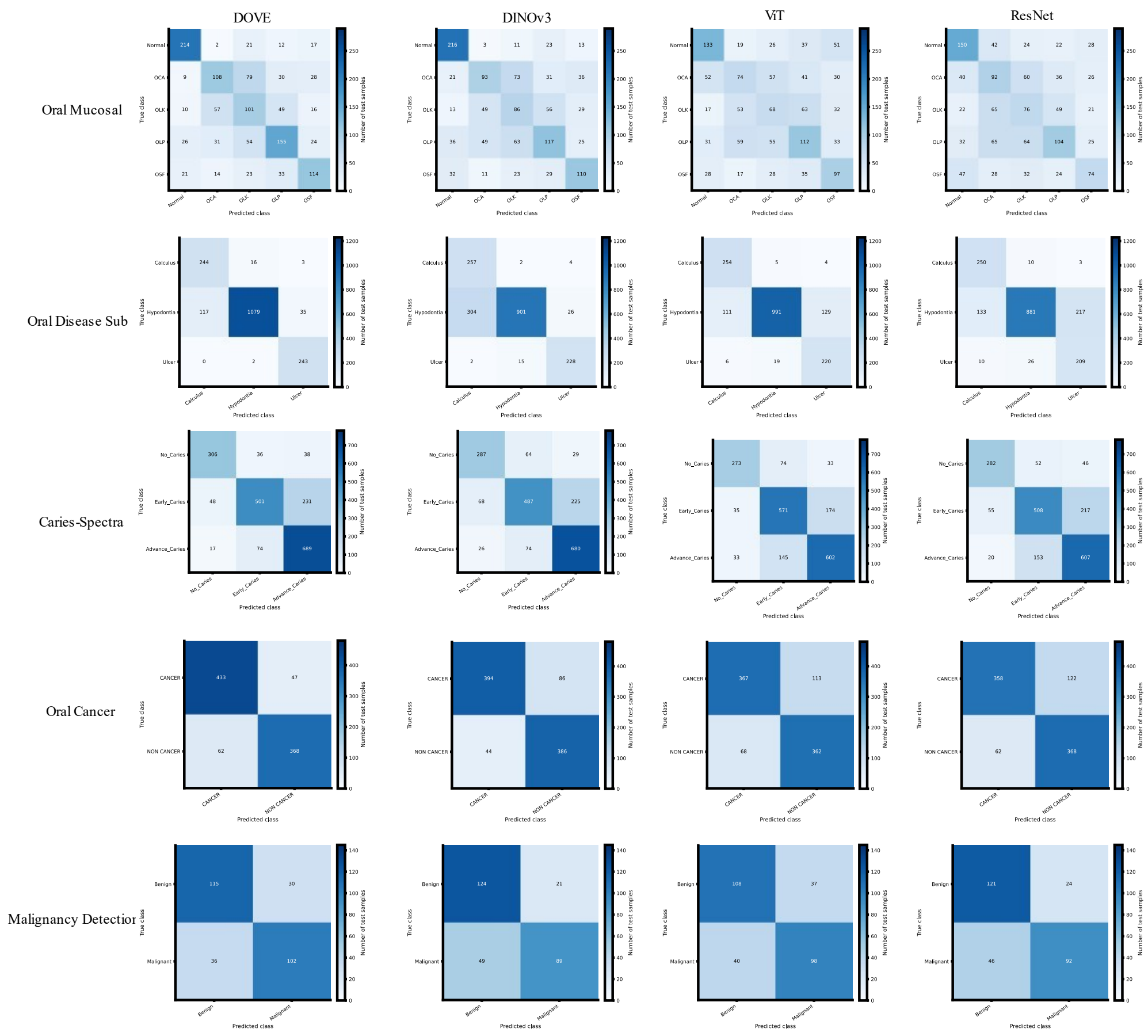

**Figure 3.** Confusion matrices of DOVE and other three baselines across five classification datasets.

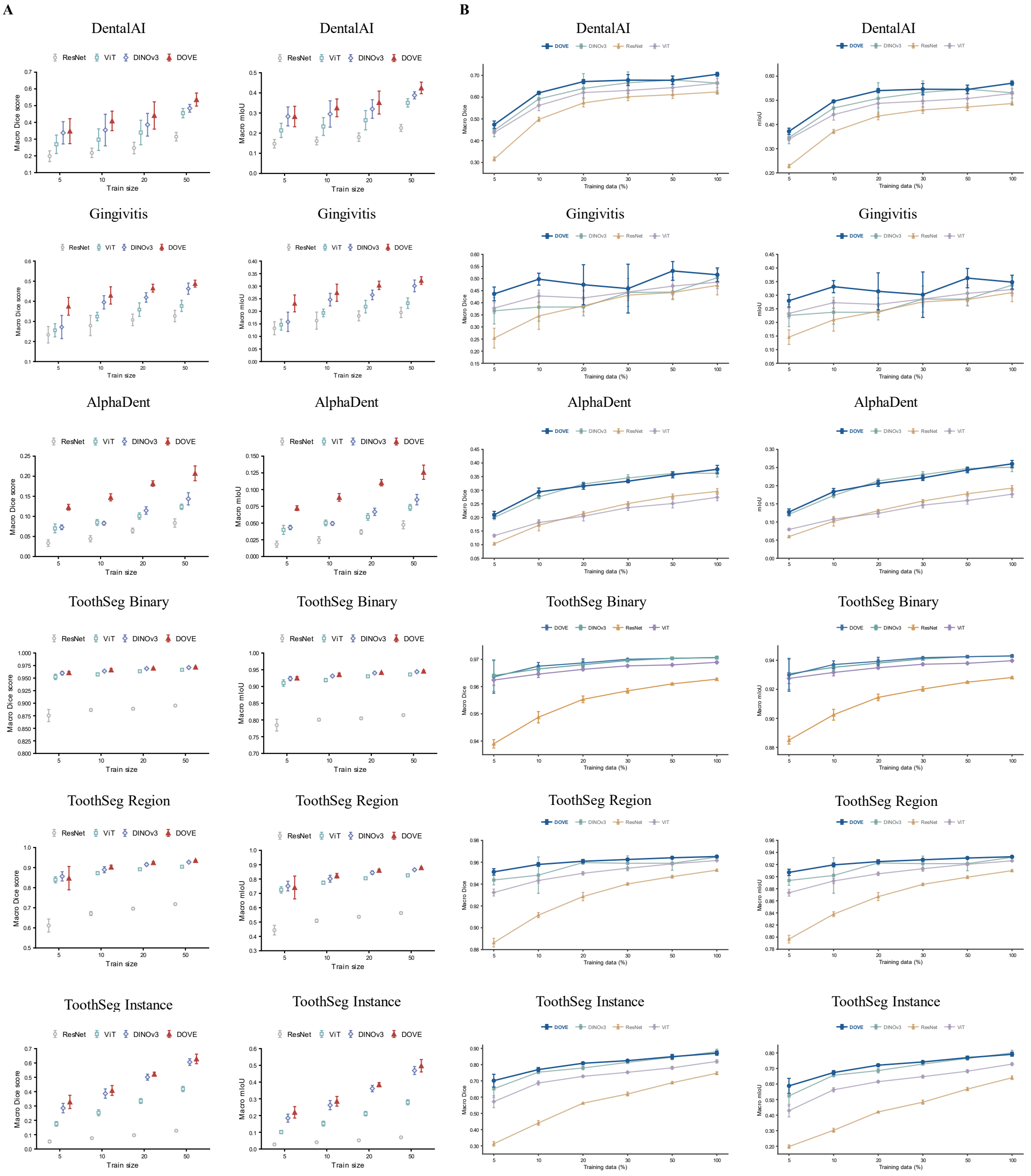

**Figure 4.** Segmentation performance. (A) Frozen backbone small-sample transfer, evaluated using a linear segmentation probe trained with 5, 10, 20, or 50 labeled images. (B) End-to-end finetuning using 5%, 10%, 20%, 30%, 50%, or 100% of the training set. Both panels report macro Dice and mIoU for DOVE and the 3 baselines. Error bars indicate standard deviations across 5 repeated runs.

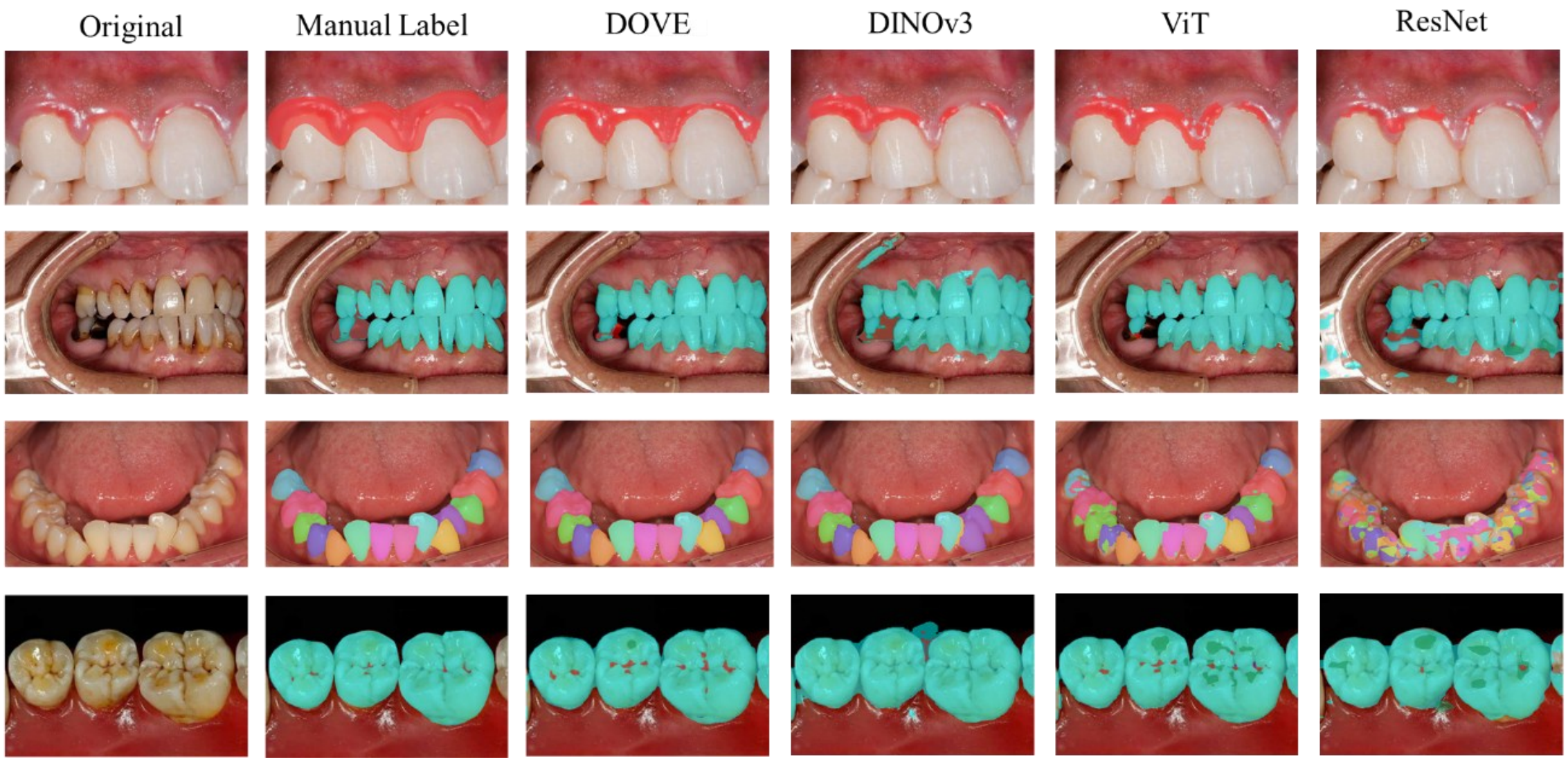

**Figure 5.** The segmentation predictions generated by trained DOVE and baselines with the manual masks.

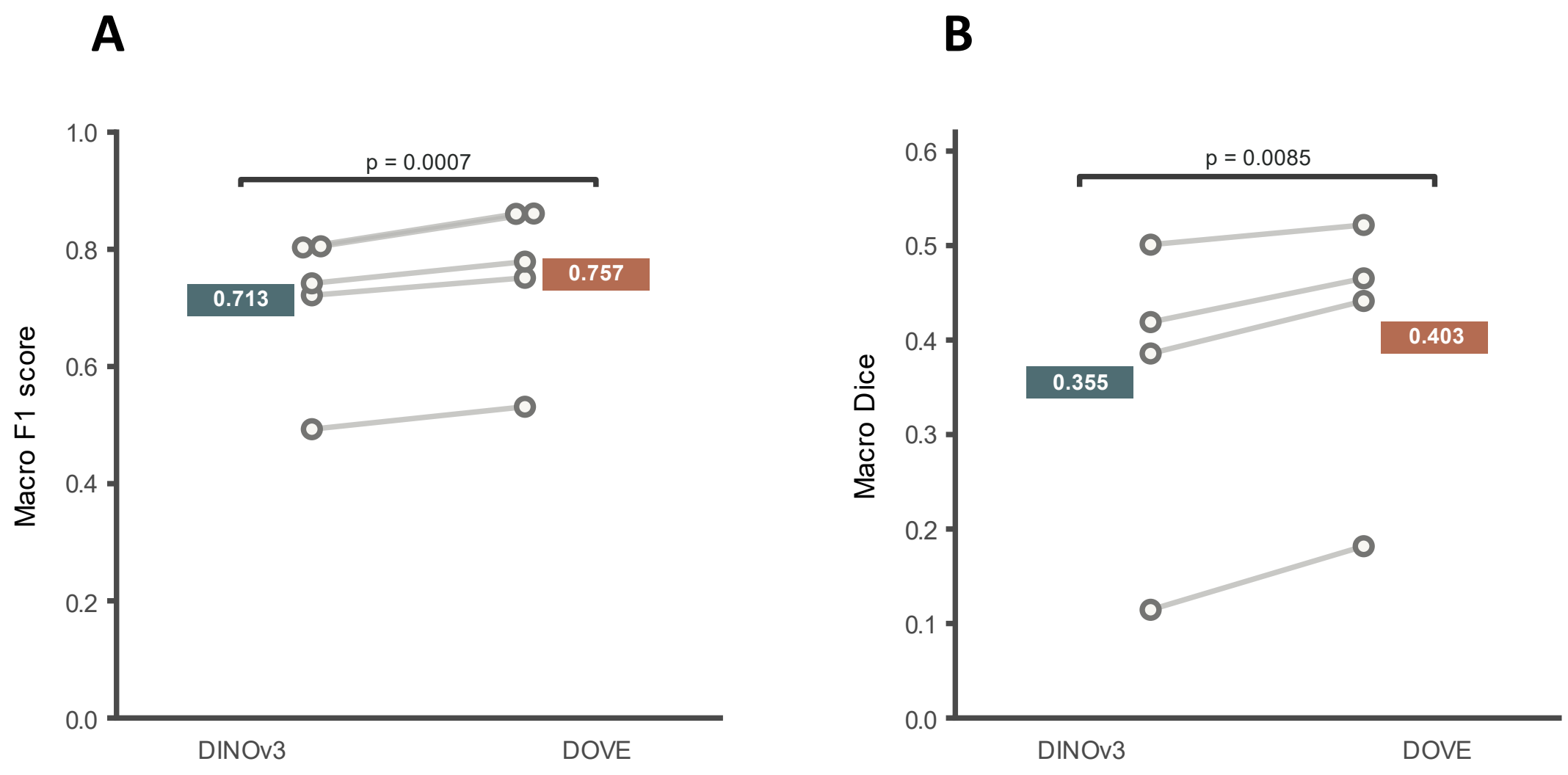

**Figure 6.** Overall performance comparisons between DOVE and DINOv3. (A) Classification performance measured by macro-F1. (B) Segmentation performance measured by macro-Dice. Each open circle represents the mean for one dataset across five repeats, and the gray lines connect the results for the same dataset under the two models. The colored floating bars indicate the average across datasets.

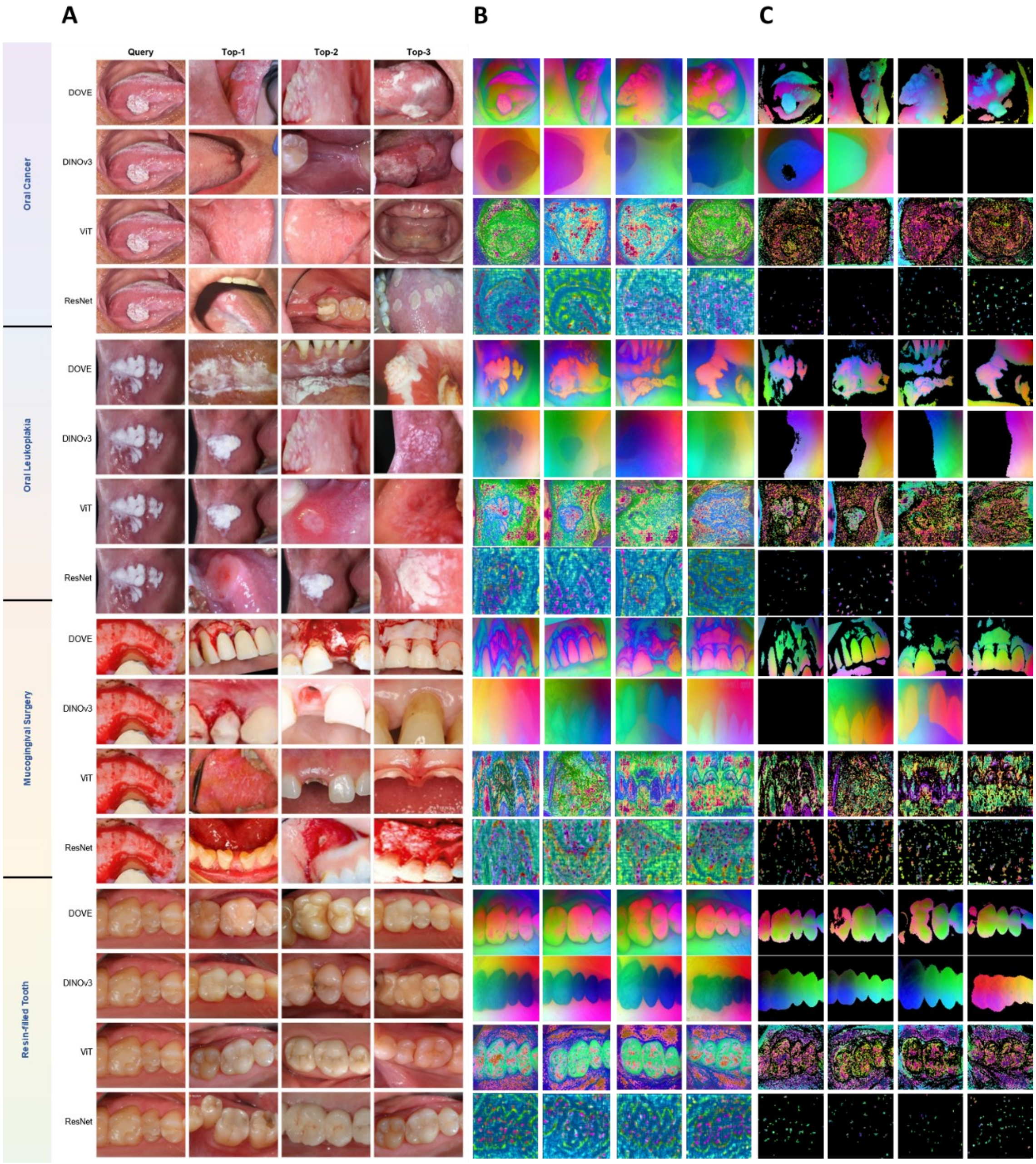

**Figure 7.** Image retrieval results and local-feature principal component analysis (PCA) visualizations across visual backbones. (A) Image retrieval results for 4 representative query cases. Each row shows the query image followed by the 3 highest-ranked results according to cosine similarity (Top-1 to Top-3). (B) PCA visualization of local features extracted from each query image and its 3 retrieved images. DOVE generates distinct semantic embeddings that clearly separate lesion tissue from surrounding mucosa, while the representations learned by baseline encoders are substantially less discriminative. (C) The background was removed by Otsu thresholding of the first principal component. In these representative examples, DOVE produced clearer and more spatially coherent foreground–background separation than the baseline encoders.

# Tables

**Table 1.** Downstream Evaluation Datasets.

| Dataset | Source | Task | Classes | Number | Region |
|---|---|---|---|---|---|
| **Oral Cancer Dataset [36]** | Public | Classification | 2 | 950 | Not reported |
| **Caries-Spectra [35]** | Public | Classification | 3 | 2,000 | Bangladesh |
| **Oral Lesions: Malignancy Detection Dataset [34]** | Public | Classification | 2 | 323 | India |
| **Oral Disease Sub [38]** | Public | Classification | 3 | 1,799 | Not reported |
| **Dataset-of-oral-mucosal-diseases [37]** | Public | Classification | 5 | 1,348 | China |
| **DentalAI [39]** | Public | Segmentation | 4 | 2,495 | Not reported |
| **AlphaDent [40]** | Public | Segmentation | 9 | 1,320 | Russia |
| **ToothSeg** | Private | Segmentation | 2/3/33 | 1,106 | China |
| **Gingivitis** | Hybrid | Segmentation | 2 | 296 | Not applicable |

# Appendix

## 1. Pretraining Corpus

The pretraining corpus was constructed by curating images from two large-scale, internet-sourced repositories: LAION [31]and MetaDent [26]. LAION contains approximately 5 billion web-crawled images, from which we initially filtered to obtain 850k dental-related candidates. MetaDent contains ~60k images derived from the COYO-700M [43], in-house datasets, and public datasets. These initial candidates underwent the curation pipeline, including de-compositing, deduplication, and content filtering to ensure relevance and quality and to remove any overlap with the downstream evaluation datasets (as described in the main manuscript). This process resulted in a final training corpus of approximately 530k images. Random samples from the dataset are shown in Appendix Figure 1.

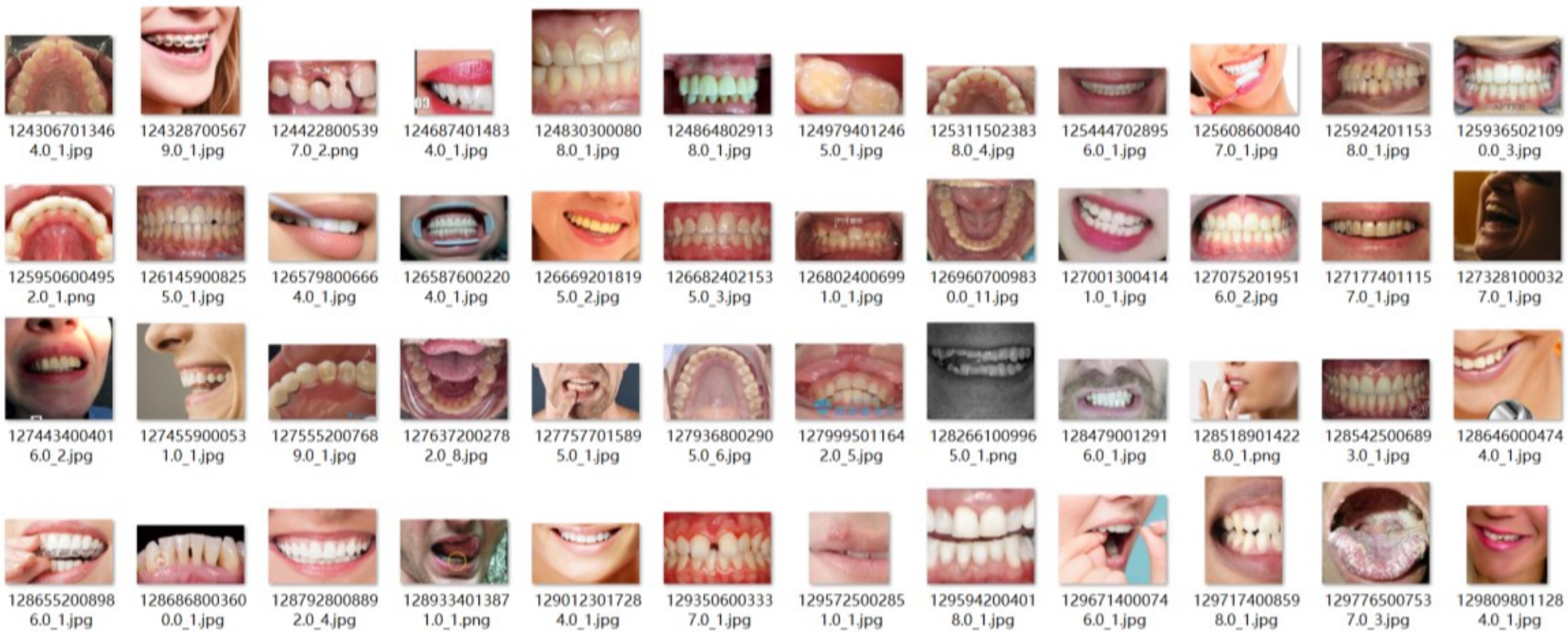

**Appendix Figure 1.** Examples from the pretraining corpus.

## 2. Assessment of Semantic Recognition

To evaluate the reliability of the semantic recognition process, 1,000 images were randomly sampled from each of three categories: excluded non-oral images, excluded facial oral images, and retained oral images. A dentist assessed the purity of the three categories. Classification precision exceeded 99% across all three categories. The high precision for non-oral and facial oral images indicates few misclassifications, whereas the high precision for retained oral images indicates a low false-retention rate. These findings indicate that the screening procedure is highly reliable in excluding privacy-sensitive and irrelevant images. The method used the vision-language model Qwen3.6-35B-A3B, and the prompts are provided in Appendix Section 6.

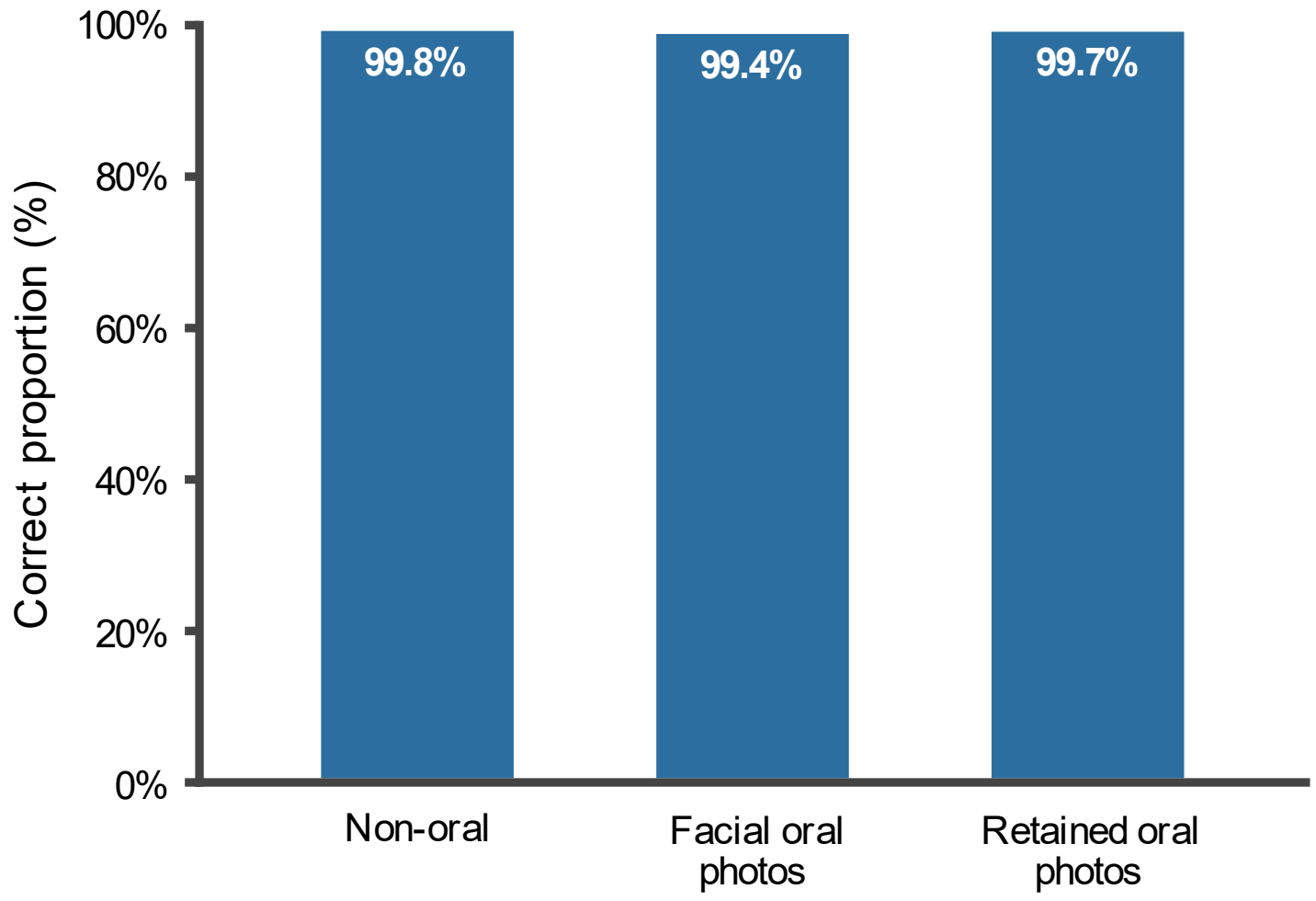

**Appendix Figure 2.** Quality assessment of semantic recognition.

## 3. Evaluation Datasets

This section provides detailed descriptions and representative images of the datasets used for downstream evaluation. The evaluation datasets encompassed classification and semantic segmentation tasks and included both publicly available and privately collected oral photographs. They covered a broad range of clinical conditions and anatomical structures, including dental caries, oral mucosal lesions, oral cancer, gingivitis, tooth abnormalities, restorations, and individual teeth. The images were collected across multiple geographic regions, including Bangladesh, India, China, and Russia, using diverse imaging devices and acquisition protocols.

**Oral Cancer Dataset** [36]

This dataset comprises 500 oral cancer images and 450 non-cancer oral images. Each image is labeled for binary classification.

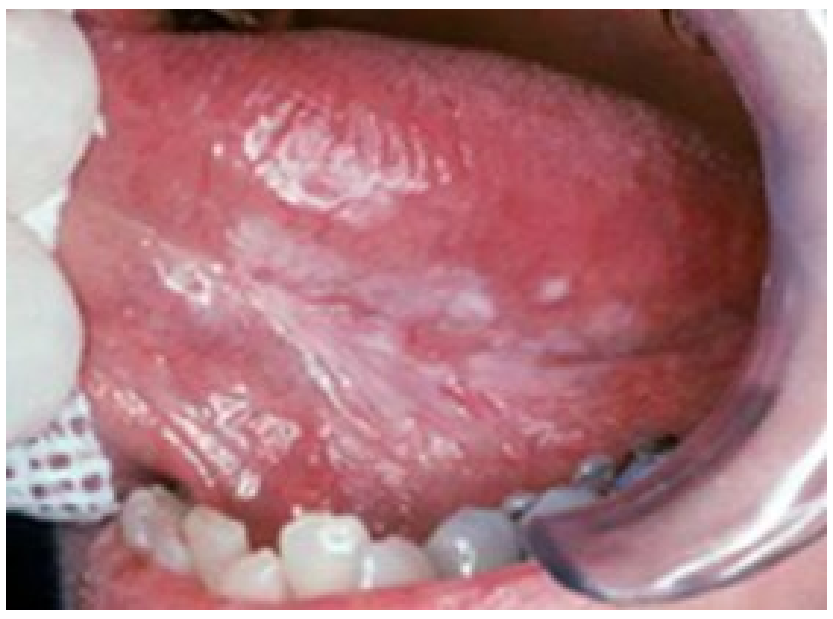

Cancer

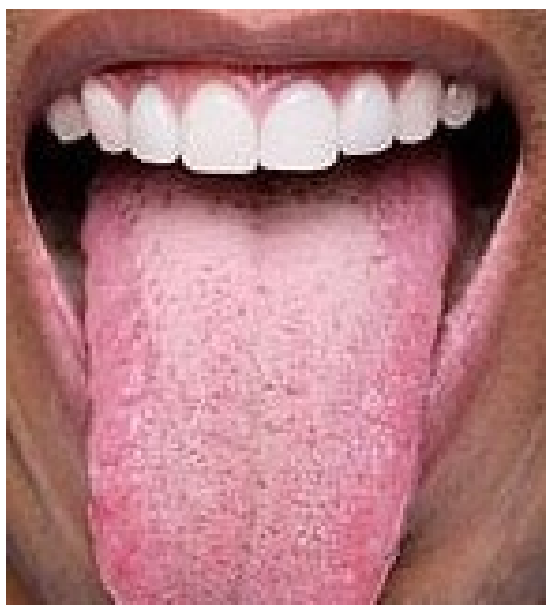

Non Cancer

**Appendix Figure 3.** Representative images of the Cancer and Non Cancer classes in the Oral Cancer Dataset.

**Caries-Spectra** [35]

This dataset comprises a diverse collection of low-resolution images featuring two types of dental caries along with a category that represents no caries. The images were sourced from several medical clinics in Rajshahi and Dhaka, Bangladesh, spanning the time period from 2021 to 2022. The dataset comprises 2,000 original JPG images across three categories: Advanced Enamel Caries (n=800), Early Stage Enamel Caries (n=800), and No Enamel Caries (n=400).

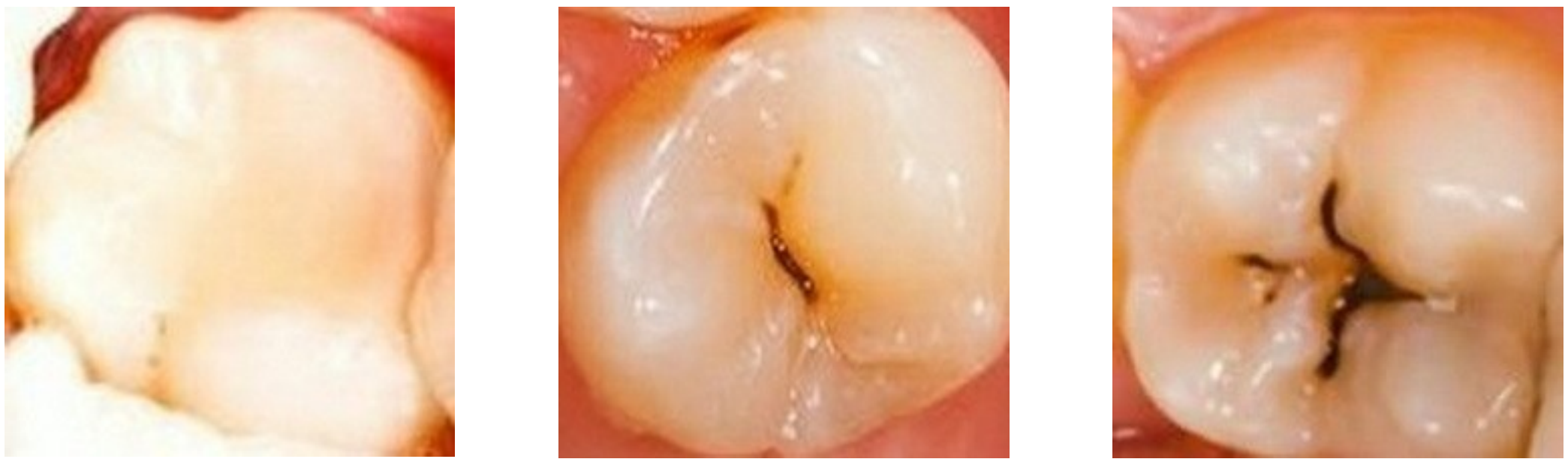

No Enamel Caries  Early Stage Enamel Caries  Advanced Enamel Caries

**Appendix Figure 4.** Representative images of the No Enamel Caries, Early Stage Enamel Caries, and Advanced Enamel Caries classes in the Caries-Spectra dataset.

**Oral Lesions: Malignancy Detection Dataset** [34]

The dataset includes color images of oral lesions captured using mobile cameras and intraoral cameras. These images can be used for identifying potential oral malignancies by image analysis. These images have been collected from different hospitals and colleges in Karnataka, India. The dataset contains images of 165 benign lesions and 158 malignant lesions.

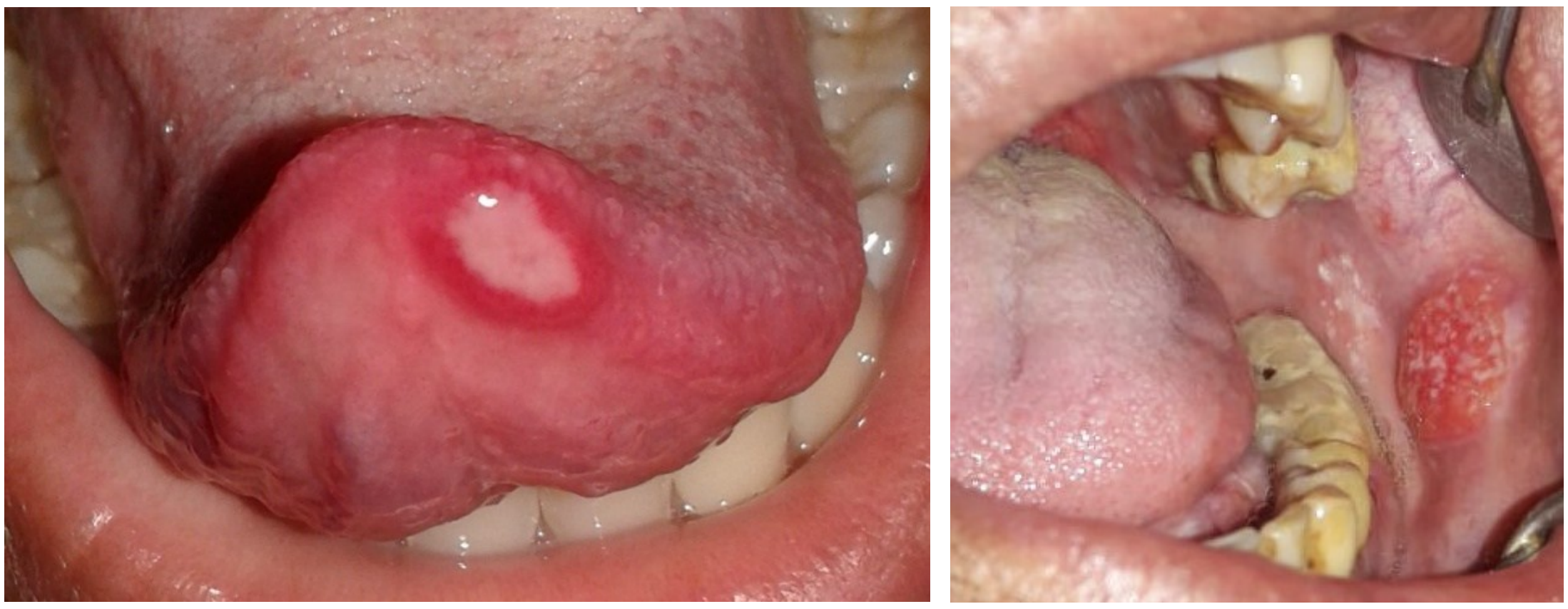

Benign Lesions  Malignant Lesions

**Appendix Figure 5.** Representative images of the Benign Lesions and Malignant Lesions classes in the Oral Lesions: Malignancy Detection Dataset.

**Oral Disease Sub**

The Oral Disease Sub Dataset is a subset of the Oral Disease [38], a comprehensive collection of images curated for dental research and analysis. Our review revealed that several categories were broadly defined and exhibited substantial semantic overlap. We therefore retained only three categories with reliable and unambiguous labels for downstream evaluation: calculus, hypodontia, and ulcer. The final dataset contains 1,799 images.

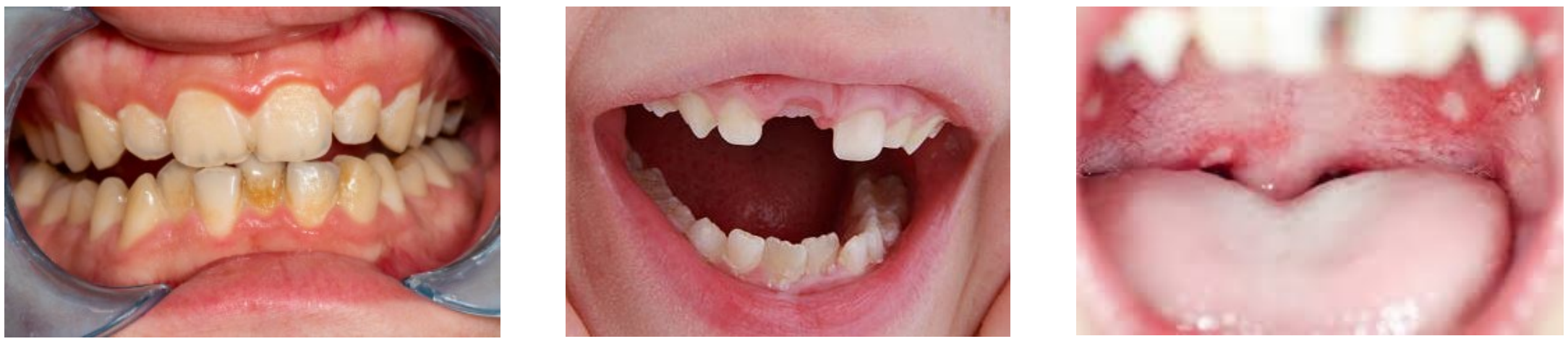

Calculus Hypodontia Ulcer

**Appendix Figure 6.** Representative images of the Calculus, Hypodontia, and Ulcer classes in the Oral Disease Sub dataset.

**Dataset-of-oral-mucosal-diseases** [37]

Most images were collected from three hospitals in China: Changsha Stomatological Hospital, the Second Xiangya Hospital of Central South University, and the First Affiliated Hospital of Hunan University of Chinese Medicine. The images were acquired using a range of smartphones and digital cameras. The dataset comprises 1,348 intraoral mucosal images categorized into five classes: normal oral mucosa, oral leukoplakia, oral lichen planus, oral submucous fibrosis, and oral cancer. The lesions were distributed across multiple oral sites, including the lips, buccal mucosa, hard and soft palate, floor of the mouth, gingiva, and anterior two-thirds of the tongue.

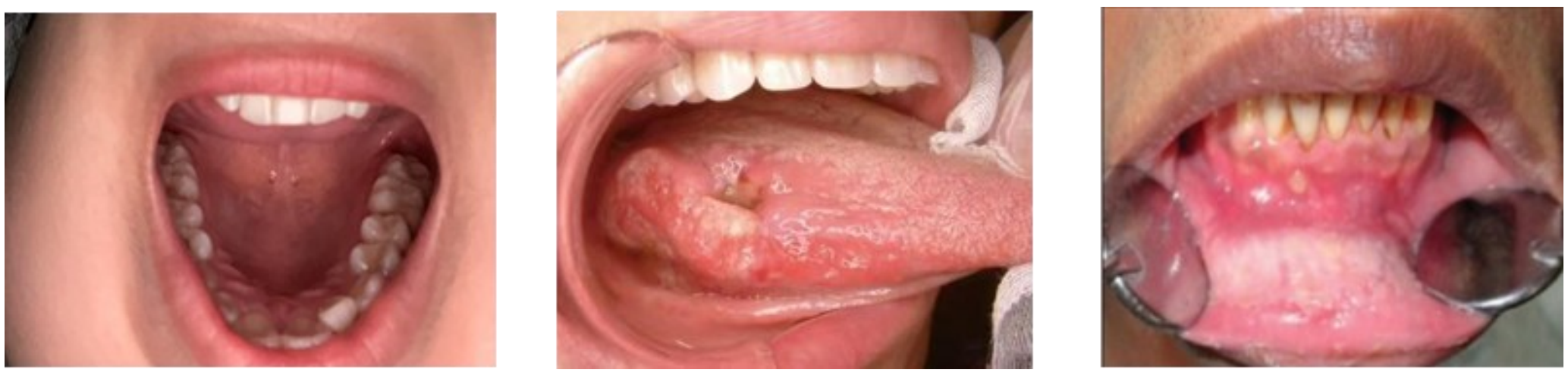

Normal Oral Cancer Oral Leukoplakia

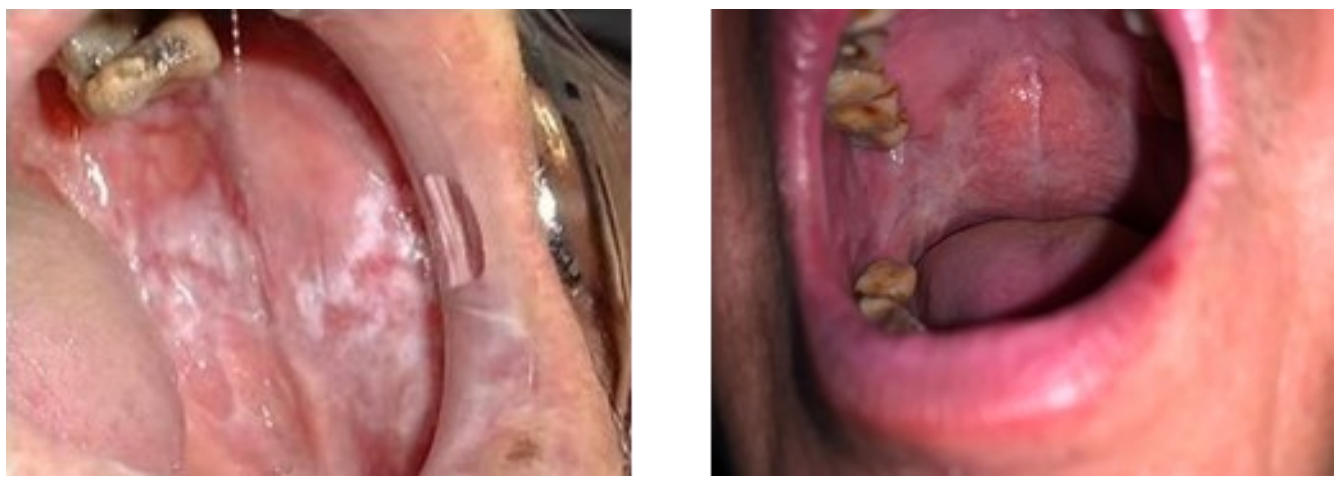

Oral Lichen Planus Oral Submucous Fibrosis

**Appendix Figure 7.** Representative images of Normal, Oral Cancer, Oral Leukoplakia, Oral Lichen Planus, and Oral Submucous Fibrosis classes in the Dataset of Oral Mucosal Diseases.

**AlphaDent** [40]

AlphaDent is an intraoral photographic dataset collected from 295 patients in Moscow, Russia. It contains 1,320 high-resolution DSLR photographs acquired using a Canon 6D Mark II camera and a 100-mm macro lens. The images include frontal and maxillary or mandibular occlusal views. Dentist-annotated instance masks are provided for nine categories: abrasion, fillings, crowns, and six classes of dental caries.

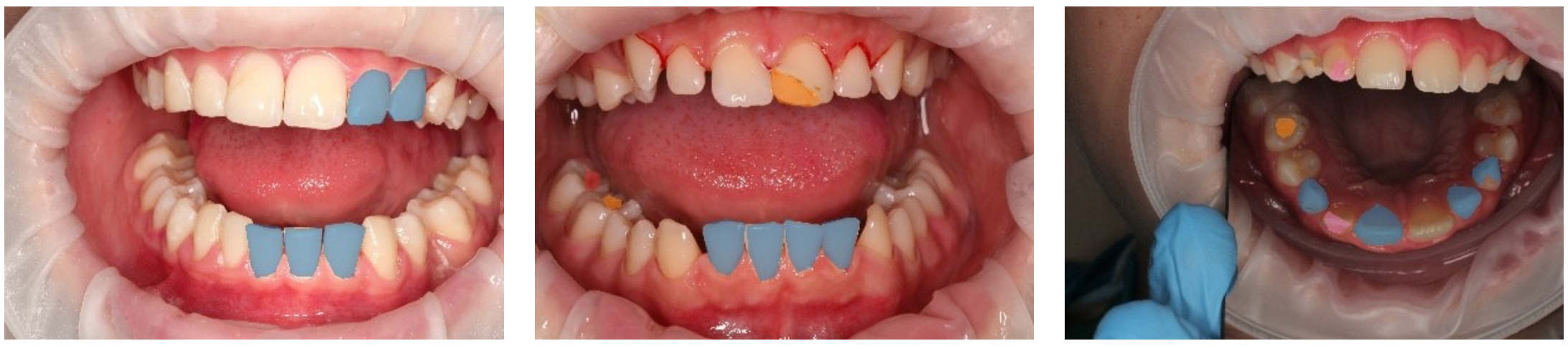

**Appendix Figure 8.** Representative images from the AlphaDent dataset.

**DentalAI** [39]

The dataset consists of 2,495 images with 28,904 labeled objects belonging to 4 different classes including tooth, caries, cavity, and other: crack. There are 3 splits in the dataset: train (1991 images), valid (254 images), and test (250 images).

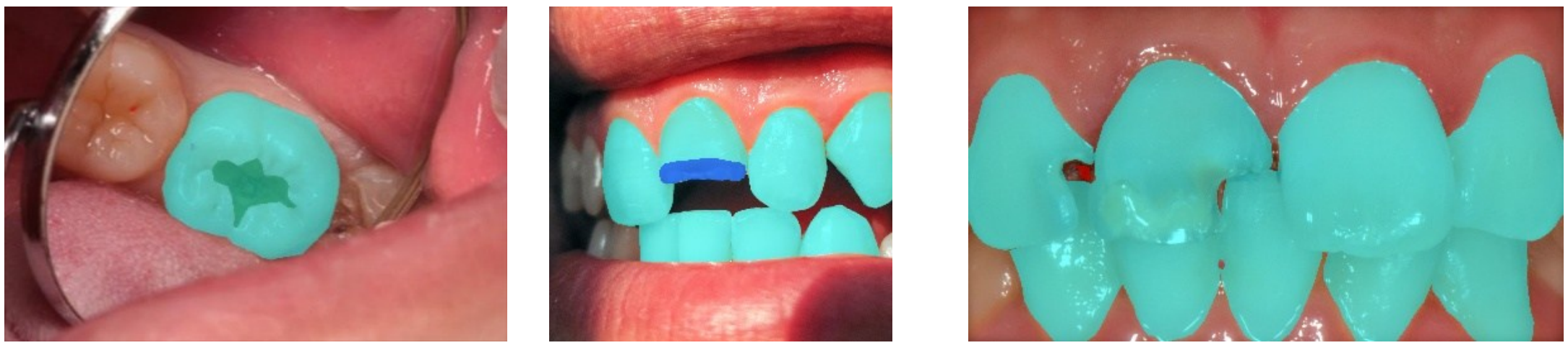

**Appendix Figure 9.** Representative images from the DentalAI dataset.

**ToothSeg**

The dataset was annotated using an in-house software platform developed by our team (https://github.com/MenxLi/labelSys). Nine intraoral photographs were acquired for each participant, and 1,106 images remained usable after excluding unusable ones. The dataset provides high-precision tooth segmentation annotations indexed by tooth position, together with labels indicating the image acquisition view.

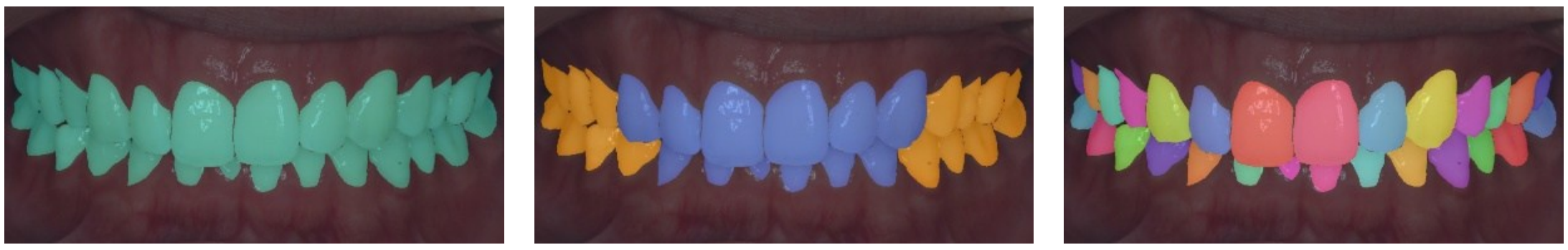

**Appendix Figure 10.** Representative images from the ToothSeg dataset. ToothSeg was evaluated at 3 levels of label granularity: binary tooth-versus-background segmentation; 3-class segmentation of anterior teeth, posterior teeth, and background; and 33-class segmentation in which each of 32 teeth was treated as a separate class, with background as an additional class.

**Gingivitis**

A gingivitis segmentation dataset was curated from MetaDent by identifying candidate images using the keywords "gingivitis," "gingival redness and swelling," "gingival swelling," "gingival redness," "periodontitis," "red gingiva," "redness and swelling of the gingival papilla," "generalized redness and swelling of the gingival papillae," "generalized gingival redness and swelling," and "extensive gingival redness and swelling". Because publicly available gingivitis segmentation datasets are limited, the existing MetaDent annotations were supplemented with additional annotations created by our team. After screening and annotation, 296 images were retained for downstream evaluation. We publicly release the dataset at https://huggingface.co/datasets/zhang-zhongshi/Gingivitis-segmentation

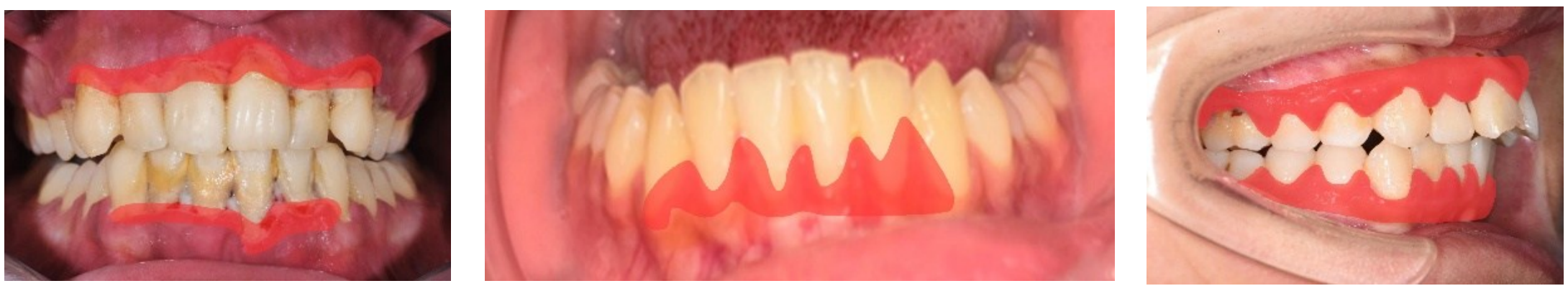

**Appendix Figure 11.** Representative images from the Gingivitis dataset.

## 4. Qualitative Analysis of DOVE Segmentation Failure Cases

To illustrate the limitations of DOVE, Appendix Figure 12 presents three representative cases of suboptimal segmentation performance. These include the misclassification of a dental cavity as caries, overestimation of the gingivitis region, and minor segmentation inaccuracies for teeth located near image boundaries.

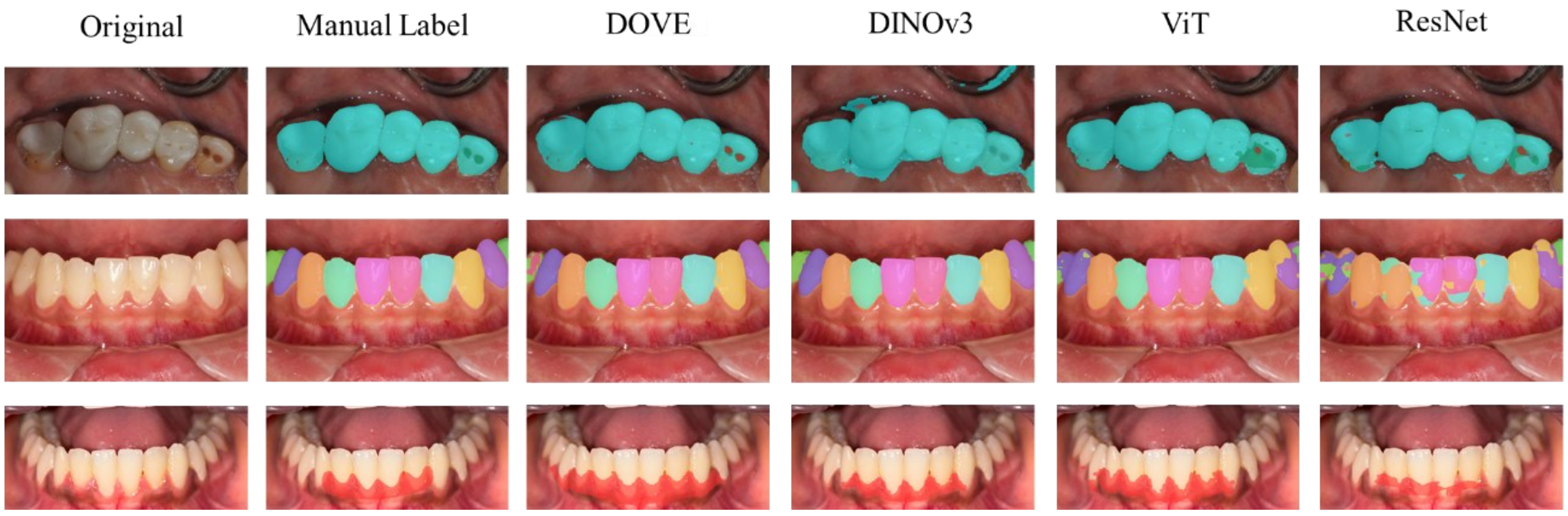

**Appendix Figure 12.** Representative segmentation failure cases of DOVE.

## 5. Evaluation Metric

### Dice Similarity Coefficient

The Dice similarity coefficient (DSC) was used to quantify the overlap between the predicted segmentation region and the ground-truth annotation. For a given class, the Dice coefficient was defined as:

$$Dice = \frac{2|P \cap G|}{|P| + |G|} = \frac{2TP}{2TP + FP + FN}$$

where P denotes the set of pixels predicted as the target class by the model, and G represents the corresponding ground-truth pixel set. TP, FP, and FN represent the numbers of true-positive, false-positive, and false-negative pixels, respectively.

The Dice coefficient ranges from 0 to 1, with higher values indicating greater overlap between the predicted and reference regions. A Dice value of 1 indicates complete agreement between prediction and ground truth, whereas a value of 0 indicates no overlap. Since Dice does not consider true-negative pixels, it is less affected by large background regions and is therefore suitable for segmentation tasks involving small foreground regions or imbalanced class distributions.

### Mean Intersection over Union

Mean intersection over union (mIoU) is a widely used metric for evaluating semantic segmentation performance by measuring the overlap between predicted regions and ground-truth annotations.
For class c, the intersection over union (IoU) is defined as:

$$IoU_c = \frac{|P_c \cap G_c|}{|P_c \cup G_c|} = \frac{TP_c}{TP_c + FP_c + FN_c}$$

Where $P_c$ denotes the set of pixels predicted as class c, and $G_c$ denotes the corresponding ground-truth pixel set. TPc, $FP_c$, and $FN_c$ represent the numbers of true-positive, false-positive, and false-negative pixels for class c, respectively.
The IoU ranges from 0 to 1, with higher values indicating better agreement between predicted and ground-truth regions. An IoU value of 1 represents perfect segmentation, whereas an IoU value of 0 indicates no overlap.
For a segmentation task containing C classes, mIoU is calculated as the arithmetic mean of IoUs across all classes.

**Accuracy**
For single-label classification tasks containing C mutually exclusive categories, each sample is assigned to only one ground-truth class. The category with the highest predicted probability is selected as the final prediction, and the resulting accuracy is referred to as accuracy or Top-1 accuracy:

$$\text{Accuracy} = \frac{1}{N}\sum_{i=1}^{N} I\,(\hat{y_i} = y_i)$$

Where N denotes the number of samples, $y_i$ and $\hat{y_i}$ represent the ground-truth and predicted labels of the i-th sample, respectively. $I(\cdot)$ denotes the indicator function, which equals 1 when the condition is satisfied and 0 otherwise.

**Macro F1-score**
Macro F1-score was used to evaluate overall classification performance across multiple categories. It is calculated by first computing the F1-score independently for each class and then averaging the values across all classes, assigning equal importance to each category.
For class c, precision and recall are defined as:

$$\text{Precision}_\text{c} = \frac{TP_c}{TP_c + FP_c}$$

$$\text{Recall}_\text{c} = \frac{TP_c}{TP_c + FN_c}$$

Where $TP_c$, $FP_c$, and $FN_c$ represent the numbers of true-positive, false-positive, and false-negative samples for class c, respectively.
The F1-score for class c is calculated as the harmonic mean of precision and recall:

$$F1_c = \frac{2Precision_c Recall_c}{Precision_c + Recall_c}$$

For a classification task containing C classes, Macro F1 is defined as:

$$Macro\ F1 = \frac{1}{C}\sum_{c=1}^{C} F1_c$$

The Macro F1-score ranges from 0 to 1. Higher values indicate better and more balanced classification performance across different categories.

**Area Under the Receiver Operating Characteristic Curve (AUROC)**

AUROC was used to evaluate the discriminative ability of classification models across different decision thresholds. It is calculated by plotting the true-positive rate (TPR) against the false-positive rate (FPR) at varying classification thresholds and measuring the area under the resulting receiver operating characteristic (ROC) curve.

For class c, the true-positive rate and false-positive rate are defined as:

$$TPR_c = \frac{TP_c}{TP_c + FN_c}$$

$$FPR_c = \frac{FP_c}{FP_c + TN_c}$$

Where TPc, FPc, TNc, and FNc represent the numbers of true-positive, false-positive, true-negative, and false-negative samples for class c, respectively.

The AUROC for class c is calculated as the area under its ROC curve:

$$AUROC_c = \int_0^1 TPR_c(x)dx$$

Where x represents the false-positive rate.

For multiclass classification, AUROC was calculated using a one-versus-rest strategy, in which each class was treated as the positive class and all remaining classes as the negative class. For a classification task containing C classes, the macro-averaged AUROC is defined as:

$$Macro\ AUROC = \frac{1}{C}\sum_{c=1}^{C} AUROC_c$$

The AUROC ranges from 0 to 1, with a value of 0.5 indicating random discrimination and a value of 1 indicating perfect discrimination. Higher values indicate better discriminative performance across different classification thresholds.

## 6. Prompts for Semantic Recognition

The following prompt was used to guide Qwen3.6-35B-A3B in identifying and excluding irrelevant or privacy-sensitive images during dataset curation. It specified the criteria for retaining oral photographs and excluding non-oral content, artificial tooth models, and images containing human eyes.

```
PROMPT = (
    """
You are a dental image screening assistant.

Your task is to classify the image based primarily on its VISUAL CONTENT.

KEEP rule
```

Set "label": "yes" if the main visual content is a genuine photograph of a real human and shows either:

1. A human face or part of the face relevant to dental or orthodontic photography, including:
- frontal face
- left/right profile
- 45-degree oblique face
- smiling face
- lips and teeth
- mouth region

2. A human intraoral view, including:
- teeth
- maxillary or mandibular occlusal view
- frontal or lateral occlusion
- lingual or palatal view
- gingiva
- oral mucosa
- tongue
- inner lips
- other photographs taken inside the human mouth

DISCARD rule

Set "label": "no" if the main visual content is NOT a genuine photograph of a real human face or human oral cavity, including:

- advertisements, posters, presentation slides, or graphic-design layouts where the dental photograph is absent or only a minor part of the image;
- collages, multi-panel images, before-and-after comparison layouts, or composite images;
- animal teeth or animal oral cavities;
- 3D digital models, CAD renders, plaster casts, dental models, dentures, or prosthetic teeth photographed alone;
- X-rays, CBCT, CT, panoramic radiographs, cephalograms, or other radiographic images;
- cartoons, illustrations, diagrams, or screenshots dominated by text;
- landscapes, objects, blank images, unrelated body parts, or other irrelevant content.

Decision priority

Use the following order:

1. First determine whether the main content is a genuine photograph of a real human face or oral cavity.

2. If yes, prefer "label": "yes" unless the image is clearly a collage, multi-panel layout, radiograph, model/render, or another explicitly excluded category.

Eye detection

Only if "label": "yes", determine "has_eyes".

Set "has_eyes": true if at least one human eye is visibly present in the image. Closed eyes still count if the eye region is clearly visible.

Set "has_eyes": false for:
- pure intraoral photographs;
- images showing only the mouth, lips, teeth, chin, or nose without any visible eye;
- all images where "label": "no".

Return only valid JSON in exactly this format:

{"label":"yes","has_eyes":true}

or

{"label":"yes","has_eyes":false}

or

{"label":"no","has_eyes":false}

Do not output any explanation or additional text.
"""
)

## 7. Checklist

We referred to the checklist of artificial intelligence in dental research to guide the reporting of this study [30]. The checklist includes 31 items covering study design, data, model development, evaluation, and reporting transparency. The completed checklist is provided below.

| Item No | Planning and Conducting | Y/N | Location in Manuscript |
|---|---|---|---|
| 1. | Study Goal | Yes | Introduction; Page 2-3 |
| 2. | Study Focus | Yes | Introduction; Page 3 |
| 3. | Data | Yes | Materials and Methods; Pages 3-5; Table 1 |
| 4. | Study Aim | Yes | Introduction; Page 2-3 |
| 5. | Reference Test | Yes | Materials and Methods; Page 4 |
| 6. | Clustering | Yes | Materials and Methods; Page 4 |
| 7. | Test Dataset | Yes | Materials and Methods; Page 4 |
| 8. | Computational resource | Yes | Materials and Methods; Page 4 |
| 9. | Comparator | Yes | Materials and Methods; Page 3 |
| | Reporting | | |
| 10. | Title | Yes | Page 1 |
| 11. | Abstract | No | Page 1; No structured abstract |
| 12. | Introduction | Yes | Pages 2-3 |
| 13. | Study Design | Yes | Page 3 |
| 14. | Data | Yes | Materials and Methods; Pages 3–5; Table 1 |
| 14 a. | Sampling | Yes | Materials and Methods; Pages 3-4 |
| 14 b. | Data Protection | Yes | Materials and Methods; Page 4/6 |
| 14 c. | Missing data | N/A | The image-level dataset contains no records with partial missing data; unusable images were entirely removed during the quality control stage (Page 3) |
| 14 d. | Data Processing | Yes | Materials and Methods; Pages 3\6 |
| 15. | Reference Test | Yes | Materials and Methods; Pages 4 |
| 16. | Sample Size | Yes | Materials and Methods; Page 4-5; Table 1 |
| 17. | Model | Yes | Materials and Methods; Page 3-5 |
| 17 a. | Model parameters | Yes | Materials and Methods; Page 3-4 |
| 18. | Training | Yes | Materials and Methods; Page 4-5 |
| 19. | Justify the best-performing model | Yes | Materials and Methods; Pages 4-5 |
| 20. | Evaluation | Yes | Materials and Methods; Pages 4-6 |
| 21. | Uncertainty | Yes | Results; Pages 7-8 |
| 22. | Explainability | Yes | Results; Page 7-8; Figure |
| | | | 5/7 |
| 23. | Results | Yes | Pages 6-8 |
| 23 a. | Performance Metrics and Data Partitions | Yes | Results; Pages 7-8 |
| 24. | Discussion | Yes | Pages 8-9 |
| 25. | Other | Yes | Page 9 |